%% file: main.tex
\documentclass[sigconf]{acmart}
\AtBeginDocument{%
  }

\copyrightyear{2024}
\acmYear{2024}
\setcopyright{rightsretained}
\acmConference[CIKM '24]{Proceedings of the 33rd ACM International Conference on Information and Knowledge Management}{October 21--25, 2024}{Boise, ID, USA}
\acmBooktitle{Proceedings of the 33rd ACM International Conference on Information and Knowledge Management (CIKM '24), October 21--25, 2024, Boise, ID, USA}
\acmDOI{10.1145/3627673.3679609}
\acmISBN{979-8-4007-0436-9/24/10}

\usepackage{soul}
\usepackage{url}
\usepackage[utf8]{inputenc}
\usepackage{graphicx}
\usepackage{amsmath}
\usepackage{amsthm}
\usepackage{algorithm}
\usepackage{algorithmic}
\usepackage{subcaption}
\usepackage{multirow}
\usepackage{hyperref}

\begin{document}

\title{Self-supervised One-Stage Learning for RF-based Multi-Person Pose Estimation}

\author{Seunghwan Shin}
\affiliation{%
    \institution{Vueron Technology}
    \city{Gangnam, Seoul}
    \country{South Korea}
    \orcid{0000-0003-2247-4584}
}
\email{sshnan7@gmail.com}
\date{April 2024}

\author{Yusung Kim}
\affiliation{%
    \institution{Sungkyunkwan University}
    \city{Suwon, Gyeonggi-do}
    \orcid{0000-0002-9306-8738}
    \country{South Korea}}
\email{yskim525@skku.edu}
\date{April 2024}
\authornote{Corresponding author}

\renewcommand{\shortauthors}{Shin et al.}

\begin{abstract}
In the field of Multi-Person Pose Estimation (MPPE), Radio Frequency (RF)-based methods can operate effectively regardless of lighting conditions and obscured line-of-sight situations. Existing RF-based MPPE methods typically involve either 1) converting RF signals into heatmap images through complex preprocessing, or 2) applying a deep embedding network directly to raw RF signals. The first approach, while delivering decent performance, is computationally intensive and time-consuming. The second method, though simpler in preprocessing, results in lower MPPE accuracy and generalization performance. This paper proposes an efficient and lightweight one-stage MPPE model based on raw RF signals. By sub-grouping RF signals and embedding them using a shared single-layer CNN followed by multi-head attention, this model outperforms previous methods that embed all signals at once through a large and deep CNN. Additionally, we propose a new self-supervised learning (SSL) method that takes inputs from both one unmasked subgroup and the remaining masked subgroups to predict the latent representations of the masked data. Empirical results demonstrate that our model improves MPPE accuracy by up to 15 in PCKh@0.5 compared to previous methods using raw RF signals. Especially, the proposed SSL method has shown to significantly enhance performance improvements when placed in new locations or in front of obstacles at RF antennas, contributing to greater performance gains as the number of people increases. Our code and dataset is open at Github. \href{https://github.com/sshnan7/SOSPE}{https://github.com/sshnan7/SOSPE} .
\end{abstract}



\maketitle

\input {1.introduction.tex}

\input {2.relatedwork.tex}

\input {3.Methods.tex}

\input {4.data.tex}
\input {5.training.tex}

\input {6.experiments.tex}

\input {7.Conclusion.tex}

\section{Acknowledgement} This work was partly supported by the National Research Foundation of Korea(NRF) grant funded by the Korea government (MSIT) (No. RS-2024-00348376) and the Institute of Information \& Communications Technology Planning \& Evaluation (IITP) grant funded by the Korea government (MSIT) (No. RS-2024-00438686, Development of software reliability improvement technology through identification of abnormal open sources and automatic application of DevSecOps), (No. 2022-0-01045, Self-directed Multi-Modal Intelligence for solving unknown open domain problems), (No. RS-2022-II220688, AI Platform to Fully Adapt and Reflect Privacy-Policy Changes), and (No. 2019-0-00421, Artificial Intelligence Graduate School Program).


\newpage
\bibliographystyle{ACM-Reference-Format}

\balance{}

\bibliography{acmart}

\end{document}

%% file: 1.introduction.tex
\section{Introduction}\label{sec:intro}
Multi-person pose estimation (MPPE) is one of the major computer vision tasks where the goal is to localize individuals and detect their body keypoints. Along with the development of image encoders such as convolutional neural network (CNN) \citep{simonyan2014very,he2016deep,liu2022convnet} and vision transformer \citep{dosovitskiy2020image,liu2021swin}, image-based pose estimation has also developed. However, image-based pose estimation has limitations in cases where the field of view is obstructed or lightning conditions are poor. To solve these problems, RF-based pose estimation approaches emerged as an alternative. The reflected RF signals can be used to extract position information about people and objects. RF signals can operate effectively irrespective of lighting conditions and in obscured line-of-sight situations thanks to their penetrative nature \citep{zhao2018through}. It makes them advantageous in situations where visibility is limited, for military, security, and disaster relief applications.

\begin{figure*}[!t]
\centering
\includegraphics[trim= 0 50 50 60, clip, width=0.95\textwidth]{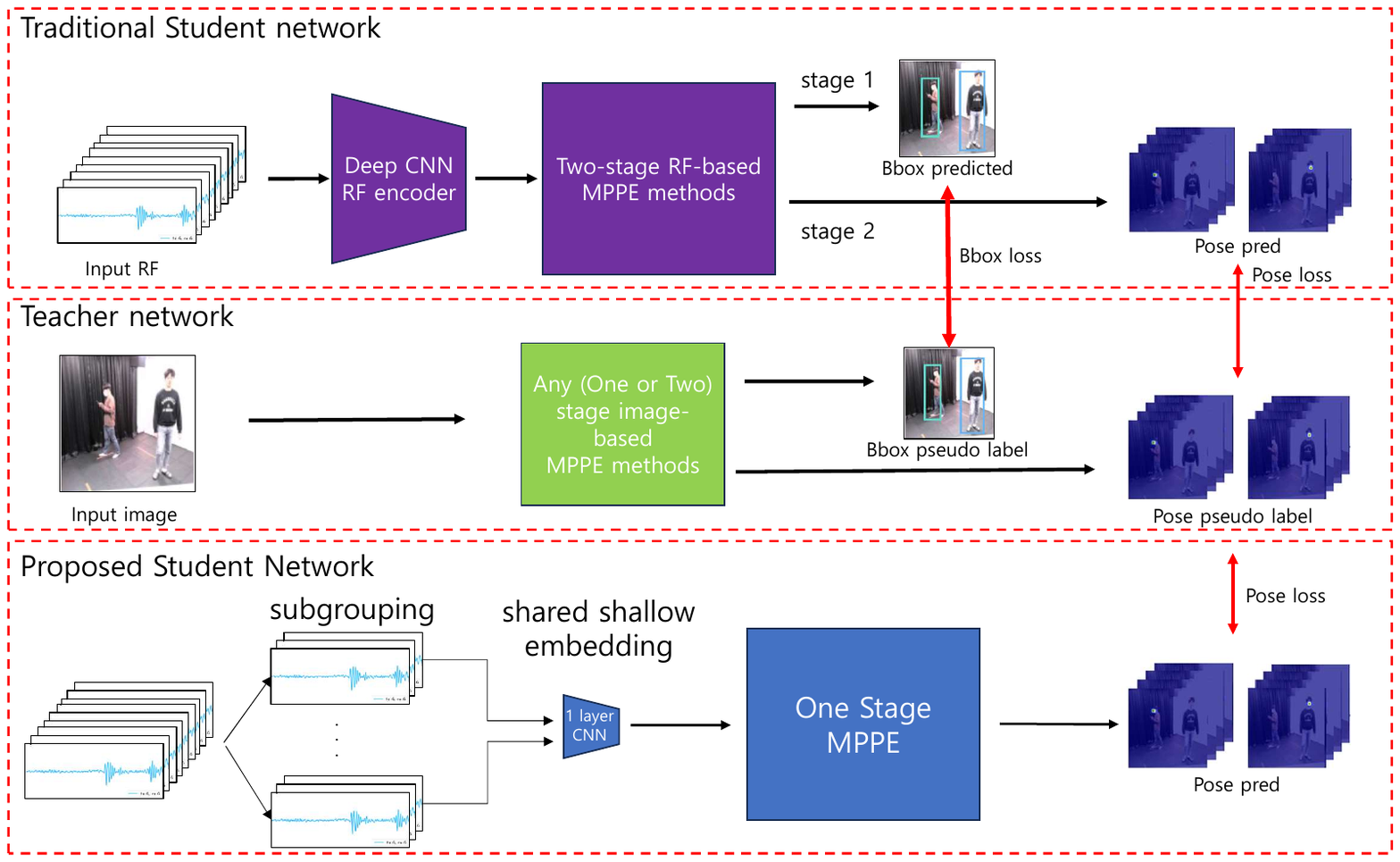}
\caption{Traditional cross-modal supervision for RF-based MPPE. Traditional RF-based MPPE used deep CNN RF encoder and two-stage MPPE methods, so they both need bounding box and keypoints label for train.}
\label{fig-cross_modal}
\end{figure*}

However, despite these advantages, RF-based pose estimation faces several challenges. The first challenge is the lack of standardized datasets or benchmarks for RF-based MPPE, unlike image-based MPPE research. For instance, image-based methods can compare the performance of proposed models simply by applying basic preprocessing (e.g., resizing) to images captured with a single camera. However, RF-based methods often use specially fabricated radar hardware ~\citep{zhao2018through, zheng2022recovering}. Some studies require strategic placement of antennas in space \citep{wang2019person}. Given the differences in wireless communication technologies, frequency bands used, the number of antennas, types, and placement strategies, the quality and data distribution of the received signals vary significantly, making direct performance comparisons difficult \citep{kim2024learning}.

The second challenge is that RF-based methods typically perform worse than image-based methods due to the difficulty of extracting accurate 2D multi-person poses from multiple 1D RF signals. Image data can directly express visual relationship information of human body joints. On the other hand, RF signals are composed of waves reflected in all directions. Traditional RF-based MPPE methods primarily involve either 1) converting RF signals into heatmap images through complex preprocessing \citep{adib20143d, zhao2018through, zheng2022recovering}, or 2) directly applying a deep embedding network to raw RF signals \citep{wang2019person, kim2024learning}. The first method delivers decent performance but requires deep knowledge of the RF domain and is computationally intensive and time-consuming. It also requires calibration depending on environmental changes. The second method, although simpler in preprocessing, requires extraction of good features through larger and deeper neural networks and has generally shown lower MPPE accuracy and generalization performance \citep{wang2019person}.

The third challenge is that existing RF-based methods rely on a two-stage framework using top-down or bottom-up approaches, making end-to-end optimization difficult. Recently, image-based methods have demonstrated superior end-to-end optimization through a one-stage end-to-end learning process, which does not require heuristic and manual post-processing such as region-of-interest (ROI) cropping \citep{he2017mask} and Non-Maximum Suppression (NMS) \citep{cao2017realtime, newell2017associative, bodla2017soft}. Image-based models cannot be directly applied to RF-based MPPE because images and RF signals inherently possess different characteristics. A delicate one-stage model design is required for RF-based MPPE, which can utilize patterns of multipath fading, signal attenuation, and reflections among RF signals received from multiple antennas.

This paper proposes an efficient and lightweight one-stage MPPE model using raw RF signals. The entire process from representing RF signals to performing MPPE requires no human intervention and optimizes end-to-end learning. 
Through encoding with self-attention mechanism with subgroup embedding, this model outperforms previous methods that encoded all signals at once through a large and deep CNN with only 2\% of the learning parameters. Additionally, this distinctive embedding architecture facilitates the design of a novel self-supervised learning method that takes inputs from one unmasked subgroup and some of the remaining subgroups masked, and then predicts the latent representations of the masked data. Empirical results demonstrate that our model improves MPPE accuracy by up to 15 in PCKh@0.5 compared to previous methods using raw RF signals. Especially, the proposed SSL method significantly enhances performance improvements in new locations or in environments with obstacles in front of RF antennas, contributing to greater performance gains as the number of people increases.

In summary, our contributions are as follows.
\begin{itemize}
\item{One-Stage Model Structure: We introduce a lightweight one-stage model structure for RF-based MPPE, enabling fully end-to-end optimization. Compared to prior work, our model has only 2\% of the number of parameters.}
\item{Signal Encoding Method: This paper presents an innovative way of embedding a subgroup of signals with a single CNN layer and subsequently integrating these subgroups using multi-head attention, improving MPPE performance.}
\item{Self-Supervised Learning : Our work introduces a novel self-supervised learning strategy that significantly enhances the representation of RF signals. This improves the overall accuracy and generalization performance of MPPE.}
\item{Open Source and Dataset: Unlike previous RF-based MPPE studies, which were predominantly closed, source codes and datasets used in this work are openly available at the provided GitHub link. This openness is intended to facilitate further research in this field by enabling other researchers.} 
\end{itemize}

%% file: 2.relatedwork.tex
\begin{figure*}[t]
\centering
\includegraphics[trim= 0 100 0 120, clip, width=1.0\textwidth, height = 0.9\columnwidth]{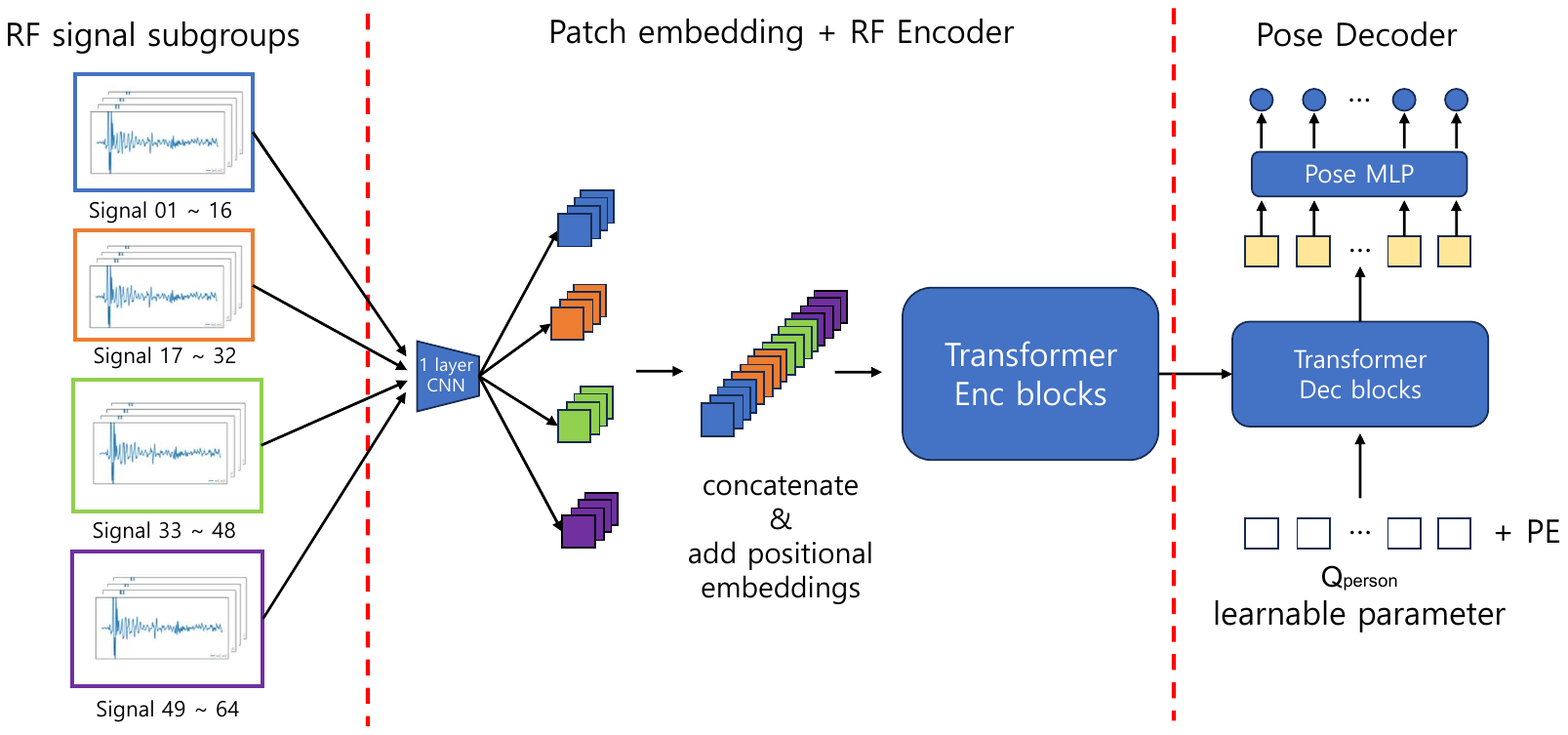}
\caption{Our model is designed with three main components for processing RF signals. Initially, the signals are categorized into subgroups, each of which is embedded into patches through a single-layer CNN. These embedded patches from all subgroups are concatenated to feed into the Encoder which extracts RF features. Finally, the Pose Decoder performs cross-attention mechanisms between the RF features and learnable query tokens $Q_{person}$ to estimate the poses of persons.}
\label{fig-overall_arch}
\end{figure*}
\section{Related works}
\subsection{Image-based multi-person pose estimation}
MPPE can be categorized into one-stage and two-stage methods. The criterion for distinguishing these methods is based on the separation of the person detection stage and the pose estimation stage. Before the advent of transformer \citep{vaswani2017attention}, performing MPPE with a one-stage method using a CNN-based decoder was challenging. After the adoption of transformers in vision tasks, as seen in models like ViT, Swin Transformer, and DETR, the concept of query-based object detection has facilitated the implementation of one-stage methods.
\newline
\newline
\textbf{Two-stage multi-person pose estimation}

MPPE which involves person detection and pose estimation as separate stages is called two-stage MPPE. Typically, the person detection stage is performed using deep learning and hand-crafted algorithms together. As a result, the performance of two-stage MPPE heavily depends on the accuracy of the person detection algorithm. If the system performs person detection first, it is called the top-down method \citep{li2021pose,sun2019deep,xiao2018simple,xu2022vitpose}. If the system performs pose estimation first, it is called a bottom-up method \citep{cao2017realtime,newell2017associative,cheng2020higherhrnet}. Since the top-down method accomplishes MPPE through iterative cropping and single-person pose estimation, it entails a longer processing time and higher computational cost compared to the bottom-up method but shows better performance. The bottom-up method incurs lower computational costs and requires less time, allowing for real-time inference but shows lower performance than the top-down method.
\newline
\newline
\textbf{One-stage multi-person pose estimation}

MPPE which simultaneously performs person detection and pose estimation is called one-stage MPPE. Utilizing transformer queries, a query-based object detection model \citep{carion2020end} is introduced. With the introduction of this query-based concept, studies on one-stage MPPE \citep{shi2022end,liu2023group} have emerged. Considering each query as a human candidate, the person detection stage became unnecessary.

\subsection{RF-based multi-person pose estimation}
Compared to image-based MPPE, there are fewer studies on RF-based MPPE, and currently, there is no RF-based one-stage method. As a two-stage method study, The RF-pose \citep{zhao2018through} demonstrated the feasibility of MPPE using RF. Additionally, \citep{zheng2022recovering} extended this concept to recover human semantic segmentation shapes. However, it's important to note that these studies employed complex RF preprocessing techniques. Person-in-WiFi \citep{wang2019person} showed that MPPE is possible even with classic commercial RF hardware. In this approach, the bottom-up method known as part affinity field \citep{cao2017realtime} was employed in the person detection stage but hand-craft methodology was still used. RPET \citep{kim2024learning} introduced an end-to-end trainable RF-based MPPE system. However, it still adheres to the top-down method. 

Many RF-based MPPE studies used cross-modal supervision for learning. They made pseudo label by image-based MPPE model. Traditional cross-modal supervision method is illustrated in Fig ~\ref{fig-cross_modal}. In this paper, we introduce a simple one-stage RF-based MPPE model that is also end-to-end trainable. Although both bounding box and keypoints information were needed for two-stage MPPE, our one-stage MPPE model is simply trained using only keypoints information.

\begin{figure*}[!ht]
\centering
\includegraphics[trim= 30 100 0 130, clip, width=0.99\textwidth]{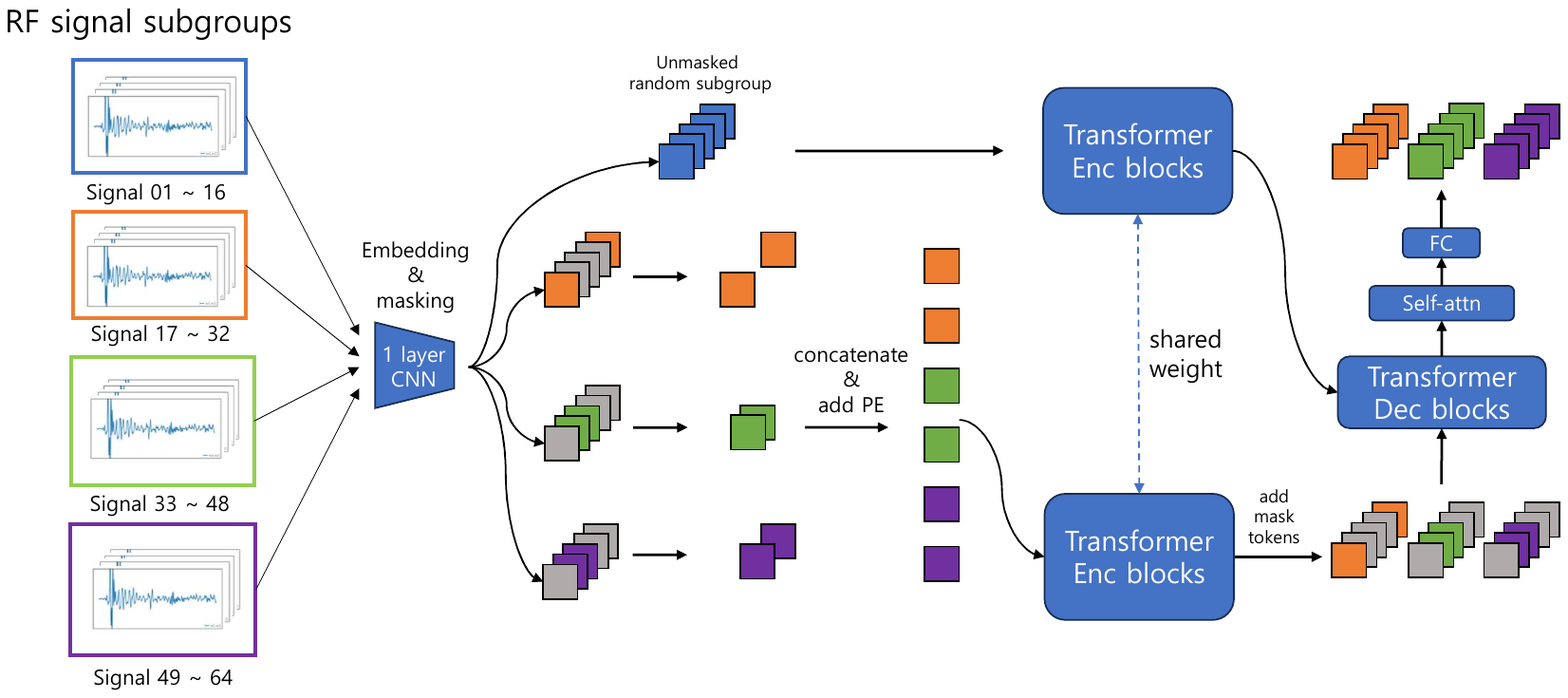}
\caption{Illustration of self-supervised learning. A random subgroup is selected, and its embedding patches are passed through the Encoder without masking to obtain the Key and Value. The patches obtained by embedding the remaining subgroups are randomly masked, and only the unmasked patches are encoded together to be used as the Query. The decoder that employs cross-attention using the obtained Key, Value, and Query is then used to reconstruct the masked patches at a latent level. The ground truth for the reconstruction is obtained by encoding all subgroups without any masking.} 
\label{fig-selfsupervsied}
\end{figure*}

\subsection{Self-supervised learning}
Manually creating labels for supervised learning is a labor-intensive task. Tasks such as extracting features from data, restoring them to the original data, or ensuring that features of augmented data resemble each other are collectively referred to as self-supervised learning, they help to alleviate the effort. Through these self-generated labels, a pre-trained model encoder can extract features more effectively. This pretrain shows superior performance even in downstream tasks. \citep{he2022masked} demonstrates that achieving higher performance in downstream tasks can be accomplished by restoring masked patches to the original image.  \citep{baevski2022data2vec,baevski2023efficient} demonstrate that restoring masked feature patches to the original features can lead to higher performance in downstream tasks. \citep{gupta2023siamese} introduced self-supervised learning in the video domain. By restoring patches from the previous frames, it achieves higher performance in downstream tasks.

%% file: 3.Methods.tex
\section{Method}

\subsection{Overview}
As part of our contributions, we present a simple straightforward model architecture. The model includes a one-layer CNN for patch embedding, a transformer encoder for RF encoding, a transformer decoder, and a multi-layer perceptron (MLP) for pose decoding. The entire training process is end-to-end trainable. The overall architecture is shown as Fig~\ref{fig-overall_arch}.
\newline
\textbf{RF encoder} \label{RF_encoder}

The RF encoder is composed of two parts: subgroup patch embedding and RF encoding. We employ a one-layer CNN for patch embedding. The detailed explanation of patch embedding will be discussed in ~\ref{ch_embedding}. As shown in Fig~\ref{fig-overall_arch}, the RF is grouped into 16 channels, and these grouped channels are independently fed into the CNN patch embedding network. So the dimension of the input RF $\mathbf{x}\in\mathbb{R}^{C\times S}$ is transformed to a patch $\mathbf{x}_{p}\in\mathbb{R}^{N\times (c\times s)}$. Where $C$ represents the total channels (64) of RF, $S$ represents the total samples (768) of RF, $c$ is the number of channels grouped, $s$ is the number of RF samples per patch and $N$ is the total number of patches.

For RF encoding, a transformer encoder layer is employed. Before feeding patches into the transformer encoder, fixed 1D positional embeddings are added to the patches. The encoder layer consists of 4 transformer encoder blocks. Through the use of a self-attention mechanism, RF signal patches are transformed into RF features $F$ that exhibit high attention scores at meaningful queries.
\newline
\newline
\textbf{Pose decoder}

The pose decoder comprises a transformer decoder, a fully connected layer (FC) for class, and a pose MLP. The RF feature $F$ and learnable parameter person queries $Q_{person}\in\mathbb{R}^{N'\times D}$ serve as inputs to the transformer decoder. Each $Q_{person}$ is a candidate for a person, where $N'$ is a fixed value of 15, and $D$ is the embedding dimension of the transformer decoder. Before being fed into the transformer decoder, positional embeddings are added exclusively to $Q_{person}$. The output of the transformer decoder, $Q_{person}'$, is then separately fed into the class FC and pose MLP. The class FC estimates whether each query represents a person (1) or not (0). The pose MLP estimates all body keypoints (both $x$ and $y$ coordinate) for each query, denoted as $P_{i}\in\mathbb{R}^{N'\times 2K}$, Where $i$ corresponds to the temporal person, and $K$ is the number of keypoints.

\subsection{Channel subgroup Patch Embedding}\label{ch_embedding}
We utilized 8 pairs of Rx and Tx antennas for data collection, where each Rx antenna can receive RF signals from all Tx antennas, resulting in a total of 64 channels of RF. Compared to image vision tasks \citep{dosovitskiy2020image, shi2022end, carion2020end}, the input channel which is 64 is too large for the first layer input. To address the problem, we conducted a process of adjusting the channels and kernel sizes of 1D CNN filters to find the optimal channel subgroup patch embedding. As a result, the RF signal $\mathbf{x}\in\mathbb{R}^{C\times S}$ transformed to $\mathbf{x}\in\mathbb{R}^{(C/N)\times N \times S}$. $N$ is the channel size of the 1D CNN filter, and $C/N$ is the number of RF subgroups formed by dividing the filter's channel. It looks similar to grouped convolution which is used in \citep{krizhevsky2012imagenet}. However, unlike grouped convolution where different networks with different weights are used for each group, we utilized a unified network to handle groups. The reason we choose unified network is RF encoder consists of a patch embedding and a transformer encoder. We first capture relative information in patch embedding part, then encodes representation vector using transformer encoder and positional encoding. We successfully applied subgroup patch embedding to the RF using the 1D CNN and further explored separation by adjusting the 1D CNN channel size. Different 1D CNN channel sizes, such as 4, 16, and 64, were experimented.

\subsection{Pretraining with Self-Supervised Learning}
For an additional experiment, we applied self-supervised learning to our dataset. The design of the self-supervised learning was specifically crafted to leverage the benefits of the channel subgroup patch embedding we introduced. The overall architecture for self-supervised learning is illustrated in Fig~\ref{fig-selfsupervsied}. The purpose of self-supervised learning in our research is to learn the RF encoder to extract features that are advantageous for pose estimation. Note that only the RF encoder part is used when fine-tuning.

As the self-supervised encoder, we employed a siamese RF encoder, where the weights are shared\citep{chen2021exploring}. This RF encoder shares the same structure as the one presented in section \ref{RF_encoder}. The RF encoder takes two inputs: one consisting of unmasked patches from a randomly selected single subgroup, and the other consisting of masked patches from the remaining subgroups. We adopted the masking method outlined in \citep{baevski2023efficient}. Each input is independently fed into the RF encoder along with positional embeddings. We manually set the mask ratio for each subgroup to 75\%. The result is two output features, denoted as $SF_{1}$ and $SF_{2}$.

The self-supervised decoder is exclusively utilized during pre-training for the reconstruction task. This decoder incorporates transformer decoder blocks, a single self-attention block, and a fully connected (FC) layer. Specifically, $SF_{1}$ is employed as the memory input for the transformer decoder, while $SF_{2}$, along with mask tokens, serves as the query input. Following the approach in \citep{he2022masked,gupta2023siamese}, positional embeddings are added to $SF_{2}$ with masked patches. After the cross-attention, an additional self-attention block is added inspired by \citep{gupta2023siamese}. The final output of each query is obtained by passing through the FC layer. 

We thought the representation feature vector should be trained because the original RF signal contains a lot of noise. To verify the assumption was correct, we defined two reconstruction labels. The first is the raw RF input signal itself, similar to \citep{he2022masked}. The second is the RF feature vector $F$, which is the unmasked output patches of the RF encoder, similar to \citep{baevski2022data2vec}. The result about self-supervised learning will be discussed in section~\ref{section-ssl}. Note that there is no gradient for the RF target vector, and the RF encoder is never trained by the target vector gradient.

%% file: 4.data.tex
\section{Data}
\subsection{Hardware}
We utilized the same dataset as introduced in \citep{kim2024learning}. Our portable radar was constructed using a commercial ultra-wideband (UWB) chip, Novelda NVA-6100 \citep{taylor:17}, which is an impulse radio ultra-wideband (IR-UWB) radar. Our radar comprises 8 $\times$ 8 antennas, and a single RF frame consists of 64 signals. The NVA-6100 transceiver has a frequency bandwidth of 0.45–3.55 GHz and achieves a spatial resolution of 7.8 mm. The signal strength is set to 30 dBm using an amplifier. For the antenna, we utilized Novelda’s Vivaldi antenna NVA-A03, which supports the 1.3–4.4 GHz frequency range and is compatible with the transceiver's bandwidth.

\subsection{Multi-Modal Data Collection} \label{section-multimodal_data}
We simultaneously collected RF signals and RGB images. The RF signal, generated by the radar, is transmitted from one of the Tx antennas and the reflected signal is received by one of the Rx antennas. We captured one frame of the RF signal set every 0.4 seconds. 

The RGB camera is securely mounted on top of our portable radar system, ensuring a consistent positional relationship between the radar antennas and the RGB camera, even when the collection location changes. Consequently, no additional calibration procedures are needed for labeling.

\begin{figure}[t]
    \centering
    \includegraphics[trim= 0 50 0 0, clip, height=0.45\linewidth, width=0.95\columnwidth]{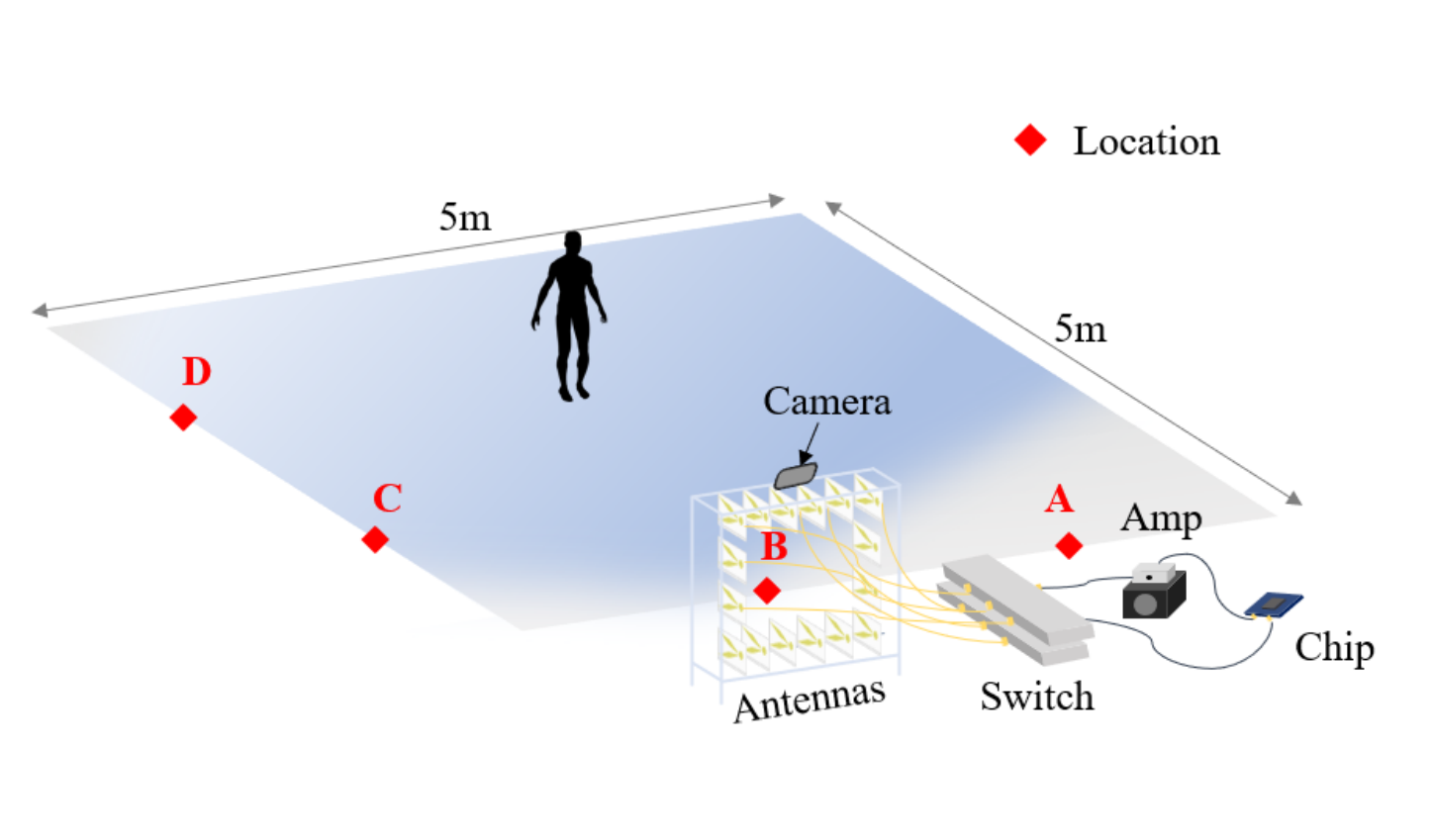}
    \caption{Data Collection Scenario: Training data was collected at four different locations—A, B, C, and D—within the same room, using a portable radar system with a fixed RGB camera.}
\label{fig:datacollect}
\end{figure}

\begin{table}[ht]
\centering
 \begin{tabular}{|c|cccc|c|} 
 \hline
 \rule{0pt}{10pt}Num. & 1 & 2 & 3 & 4 & Total \\ 
 \hline 
 \rule{0pt}{10pt}N & 87,000 & 53,000 & 24,000 & 12,000 & 176,000  \\ 
 \hline\hline
 \rule{0pt}{10pt}Loc. & A & B & C & D & Total \\ 
 \hline 
 \rule{0pt}{10pt}N & 34,000 & 33,000 & 49,000 & 60,000 & 176,000 \\
 \hline
 \end{tabular}
 \caption{Training Dataset: In a single RF frame, the maximum and average number of persons are 4 and 1.78, respectively.
\label{tab:table-dataset}}
\end{table}

We conducted training data collection at four different locations within a 5m$\times$5m room, as depicted in Fig~\ref{fig:datacollect}. During this phase, there were no obstacles between the radar system and individuals to ensure proper capture of RGB camera images for labeling. Furthermore, we tested our model in diverse scenarios: (A) changing radar locations to two different positions not in the training set within the same room, (B) placing obstacles (e.g., a piece of fabric or a foam box) in front of the radar antennas, and (C) relocating the radar to another room (untrained environment).
We gathered data from 12 volunteers engaged in daily activities such as standing, walking around the room, and stretching arms for training purposes. The number of persons present in a single RF frame ranged from 1 to 4, as outlined in Table~\ref{tab:table-dataset}. In total, 176,000 RF frames were collected for training.

\subsection{label for Pose estimation}
For supervised learning in pose estimation, we require human keypoints labels and class labels for each query in every frame. Keypoints labels consist of $x$ and $y$ coordinates of body joints. Regarding the class label, as our focus is solely on the person class, it is assigned a value of either 0 or 1, indicating the presence or absence of a person. In Section~\ref{section-multimodal_data}, we utilized the image collection and performed inference using \citep{wang2019person} to obtain pseudo keypoint labels. 

%% file: 5.training.tex
\section{Training}
\subsection{Bipartite Matching}
\citep{carion2020end} introduced query-based object detection which is end-to-end trainable. They utilized the algorithm called bipartite matching to match queries with targets by the minimum distance, enabling the model to learn without collapse. We also used bipartite matching to train our model. Calculating the distance of class and keypoints, each query is matched to the nearest label.

\subsection{Loss}
For supervised learning in pose estimation, we employ binary cross-entropy loss (BCE loss) for the class loss and mean squared error (MSE loss) for the keypoints loss. The formulas for BCE loss and MSE loss are as follows.
\begin{equation}
    \mathcal{L}_{CLS} = \frac{1}{N}\sum\limits_{i=0}^{N-1} y_{i} \cdot log \hat{y}_{i} + (1-y_{i}) \cdot log (1-\hat{y}_{i})
\end{equation}
$N$ denotes the number of queries, representing the candidates for persons, $y_{i}$ denotes class label, $\hat{y}_{i}$ denotes predicted class.
\begin{equation} \label{mseloss}
    \mathcal{L}_{pose} = \frac{1}{M}\frac{1}{K}\sum\limits_{i=0}^{M-1}\sum\limits_{k=0}^{K-1} (P_{ik} - \hat{P}_{ik})^{2}
\end{equation}
$M$ denotes the number of persons label, $P_{ik}$ denotes the k-th keypoint label of the i-th person, and $\hat{P}_{ik}$ represents the predicted k-th keypoint of the i-th person.  ${L}_{CLS}$ is computed for all query candidates, which are persons, resulting in calculations for $N$ queries. However, ${L}_{pose}$ computes the loss for the predicted keypoints that are closest in distance to the labels, based on the number of labels after bipartite matching, denoted as $M (N>=M)$. The total loss is as follows.
\begin{equation}
    \mathcal{L} = {\lambda}_{cls}{\mathcal{L}}_{cls} + {\lambda}_{pose}{\mathcal{L}}_{pose}
\end{equation}
$\lambda$ represents the weight of the loss.

For self-supervised learning, we utilize mean squared error (MSE) loss as the reconstruction loss. The formula is the same as Formula~(\ref{mseloss}).
\begin{equation} 
    \mathcal{L}_{self-supervised} = \frac{1}{T}\sum\limits_{i=0}^{T-1} (v_{i} - \hat{v}_{i})^{2}
\end{equation}
However, in self-supervised learning, $T$ represents the number of masked patches, $v_{i}$ denotes the patch vector label, and $\hat{v}_{i}$ denotes the predicted patch vector.

\begin{table}[ht]
\centering\footnotesize{
\renewcommand{\arraystretch}{1.0}
\begin{tabular}{l|l}
Configs                                                                   & Value                                                     \\ \hline
optimizer                                                                 & AdamW                                                     \\
base learning rate                                                        & 4e-4                                                      \\
weight decay                                                              & 1e-4                                                      \\
optimizer momentum                                                        & $\beta_{1}$, $\beta_{2}$ = 0.9, 0.999 \\
batch size                                                                & 64                                                        \\
learning rate schedule                                                    & cosine decay                                              \\
warmup epochs                                                             & 10                                                        \\
training epochs                                                           & 30                                                        \\ \hline
$\lambda_{cls}$, $\lambda_{pose}$ & 1, 50                                                     \\
1D CNN embedding filter shape                                             & {[}16, 8{]}                                               \\
\# of encoder blocks                                                      & 4                                                         \\
\# of embedding dimension                                                 & 128                                                       \\
\# of heads                                                               & 4                                                         \\ \hline
\# of decoder blocks                                                      & 2                                                       
\end{tabular}
}
\caption{Supervised learning setting for OS (One-Stage) MPPE}
\label{table-OS-hyperparameter}
\end{table}
\begin{table}[ht]
\centering\footnotesize{
\renewcommand{\arraystretch}{1.0}
\begin{tabular}{l|l}
Configs                       & Value                                                     \\ \hline
optimizer                     & AdamW                                                     \\
base learning rate            & 4e-5                                                      \\
weight decay                  & 5e-2                                                      \\
optimizer momentum            & $\beta_{1}$, $\beta_{2}$ = 0.9, 0.999 \\
batch size                    & 128                                                       \\
learning rate schedule        & cosine decay                                              \\
warmup epochs                 & 10                                                        \\
training epochs               & 90                                                        \\
mask ratio                    & 0.75                                                      \\ \hline
1D CNN embedding filter shape & {[}16, 8{]}                                               \\
\# of encoder blocks          & 4                                                         \\
\# of embedding dimension     & 128                                                       \\
\# of heads                   & 4                                                         \\ \hline
\# of decoder blocks          & 2                                                       
\end{tabular}
}
\caption{Pre-training setting for self-supervised learning}
\label{table-SOS-pretrain-hyperparameter}
\end{table}
\begin{table}[ht]
\centering\footnotesize{
\renewcommand{\arraystretch}{1.0}
\begin{tabular}{l|l}
Configs                       & Value                                                     \\ \hline
optimizer                     & AdamW                                                     \\
base learning rate            & 4e-4                                                      \\
weight decay                  & 5e-2                                                      \\
optimizer momentum            & $\beta_{1}$, $\beta_{2}$ = 0.9, 0.999 \\
batch size                    & 64                                                        \\
learning rate schedule        & cosine decay                                              \\
warmup epochs                 & 10                                                        \\
training epochs               & 25                                                        \\ \hline
1D CNN embedding filter shape & {[}16, 8{]}                                               \\
\# of encoder blocks          & 4                                                         \\
\# of embedding dimension     & 128                                                       \\
\# of heads                   & 4                                                         \\ \hline
\# of decoder blocks          & 2                                                       
\end{tabular}
}
\caption{Fine-tuning setting for SOS (self-supervised OS) MPPE}
\label{table-SOS-finetuning-hyperparameter}
\end{table}

\begin{table}[h]
\resizebox{0.9\columnwidth}{!}
{\small
\renewcommand{\arraystretch}{1.2}
\begin{tabular}{c|lccc}
\hline
\multirow{2}{*}{Model} & \multicolumn{4}{c}{The number of learnable parameters}                                     \\ \cline{2-5} 
                       & RF encoder & person detector & \multicolumn{1}{c|}{pose estimator} & Total \\ \hline
RPET&  37$\times$10$^6$  & 98$\times$10$^5$ &  \multicolumn{1}{c|}{47$\times$10$^6$}               &  94$\times$10$^6$     \\
SOS                    &    80$\times$10$^4$      &          -       & \multicolumn{1}{c|}{33$\times$10$^4$}               &   13$\times$10$^5$    \\ \hline
\end{tabular}
}
\caption{Comparison of the model parameter counts between RPET and SOS. We drastically reduced the model size, which increased performance.}
\label{table-parameters}
\end{table}
\section{Implementation Details}
\subsection{Hyperparameters}
In this section, we describe the key hyperparameter settings used for training the OS (One-Stage MPPE) and SOS (Self-supervised One-Stage MPPE) models. Table 2 presents the settings used for supervised learning of the OS model. Table 3 shows the pre-training settings for self-supervised learning. Table 4 displays the settings used to fine-tune the pre-trained model for MPPE.

\subsection{Number of Model Parameter}
One more contribution of our work is the lightweight model size. When comparing RPET \citep{kim2024learning} and the proposed OS model, there are two main reasons why OS has fewer model parameters. First, OS changed the encoder for raw RF signal embedding to a single CNN layer. RPET used the deeper and heavier ConvNeXT \citep{liu2022convnet} architecture. Second, with a one-stage model structure, OS does not use a separate person detection network. Table~\ref{table-parameters} shows the number of the model parameters used in RPET and OS. OS uses only 2\% of the learnable parameters compared to RPET, which achieved the previous SOTA performance, and it reduced the training time to 1/6 when trained on the same machine. In the subsequent section, it is shown that OS, with 2\% of the model parameters, achieves higher MPPE performance than RPET.

%% file: 6.experiments.tex
\section{Experiments}
The Percentage of Correct Keypoints (PCK) metric is employed to assess pose estimation accuracy, consistent with the evaluation metrics used in the MPII dataset. In RF-based pose estimation studies, \citep{wang2019person,kim2024learning} utilized PCK as an evaluation metric. Following the PCK methodology, a body joint is deemed detected correctly if the Euclidean distance between the prediction and label is smaller than a threshold based on the head size (we manually set the threshold as half of the head size, PCKh\textsubscript{50}).

To compare performance, we use the following two models trained on the same dataset: Baseline and RPET~\citep{kim2024learning}. The Baseline model is trained using the traditional Teacher-Student cross-modal supervised learning approach~\citep{kim2024learning, wang2019person}. The model is designed using a two-stage MPPE approach with a deep and large CNN embedding network. RPET is an enhanced version of the Baseline model that applies knowledge distillation to mimic the latent representations of the image-based model as the Teacher network. Additionally, we compare the performance of both models with techniques introduced in related studies for performance improvement, such as jointly learning Segmentation Masking~\citep{wang2019person}\citep{zheng2022recovering} and using a Multi-Frame version~\citep{zhao2018through} that inputs not only the current RF signal but also stacks past signals.

\subsection{Basic Performance Comparisons}
\begin{table}[]
\centering\resizebox{\columnwidth}{!}{
\renewcommand{\arraystretch}{1.3}
\begin{tabular}{l|cc|c}
\hline
Model & \# signals per subgroup & \# of subgroup & PCKh\textsubscript{50} \\ \hline
Baseline  & 64            &       1      &   74.3  \\  
Baseline \& Masking    &    64         &      1      &   75.1 (+0.8) \\
Baseline \& Multi-Frame     &    64         &      1      &   74.7 (+0.4)  \\\hline
RPET    &    64         &      1      &   78.9 (+4.6) \\
RPET \& Masking    &    64         &      1      &   78.9 (+4.6) \\
RPET \& Multi-Frame     &    64         &      1      &   79.9 (+5.6)  \\\hline
 OS      &    4        &      16     &  80.9 (+6.6)   \\
 OS     &    16         &      4      &   \textbf{83.7 (+9.4)}  \\
 OS     &    64         &      1      &   82.2 (+7.9)  \\ \hline
\end{tabular}
}
\caption{Comparison with previous works in test scenario (A) different antenna locations in the same room. The performance of OS (One-Stage MPPE) varied depending on how the 64 signals were divided into subgroups for embedding. Embedding them into four subgroups of 16 signals each achieved the best PCKh\textsubscript{50} performance.}
\label{table-Channel subgroup patch embedding result}
\end{table}

We compared the performance of MPPE with Baseline and RPET in the test scenario (A) where testing locations of the radar are changed in the same room. The results are shown in Table ~\ref{table-Channel subgroup patch embedding result}. The performance of our proposed model, OS, varied depending on how the 64 signals were divided into subgroups for embedding. Embedding the 64 signals into one subgroup resulted in lower performance compared to embedding them into four subgroups of 16 signals each, which yielded the best performance. This suggests that it is difficult to embedding and represent all 64 signals at once. Conversely, embedding the signals into 16 subgroups (4 signals each) resulted in poor performance (still higher than RPET), because each subgroup contained too few signals to extract meaningful features effectively. As a result, OS embedding to 4 subgroups empirically achieved the best PCKh\textsubscript{50} performance with our RF encoder architecture (CNN embedding and transformer encoder). 



\begin{table}[ht]
\centering
\resizebox{\columnwidth}{!}{
\renewcommand{\arraystretch}{1.2}
\begin{tabular}{c|cccccccc|c}
\noalign{\smallskip}\noalign{\smallskip}
\hline
\rule{0pt}{10pt} Model & Hea & Nec & Sho & Elb & Wri & Hip & Kne & Ank & Total \\
\hline \hline
\rule{0pt}{10pt} \begin{tabular}[c]{@{}c@{}}RPET\end{tabular} & 88.1 & 88.7 & 84.8 & 74.2 & 33.2 & 88.8 & 87.7 & 85.3 & 78.9 \\
\hline
\rule{0pt}{10pt} OS & 89.4 & 90.1 & 88.1 & 81.8 & 52.3 & 89.9 & 89.5 & 88.1 & 83.7 \\
\rule{0pt}{10pt} SOS & \textbf{89.8} & \textbf{90.4} & \textbf{88.8} & \textbf{82.5} & \textbf{55.2} & \textbf{90.5} & \textbf{90.3} & \textbf{88.8} & \textbf{84.5} \\
\hline
\end{tabular}
}
\caption{Results of each body joint PCKh\textsubscript{50} in test scenario (A) changing antenna locations in the same room. SOS (Self-supervised One Stage MPPE) denotes our proposed model with self-supervised learning.}
\label{tab:table-jointresults}
\end{table}

\begin{table}[ht]
\centering\resizebox{0.8\columnwidth}{!}{
\renewcommand{\arraystretch}{1.2}
\begin{tabular}{cc|cccl}
\cline{1-5}
\multirow{2}{*}{Test set}                                                             & \multirow{2}{*}{\# of person} & \multicolumn{3}{c}{Model}                &                      \\ \cline{3-5}
                                                                                      &                               & \multicolumn{1}{c|}{RPET} & OS & SOS &                      \\ \cline{1-5} 
\multirow{5}{*}{\begin{tabular}[c]{@{}c@{}}Different locations\\in the same room\end{tabular}}    & 1                             & \multicolumn{1}{c|}{84.8} & 90.1 & \textbf{90.9}  &                      \\
                                                                                      & 2                             & \multicolumn{1}{c|}{78.3} & 81.7 & \textbf{83.4}  &                      \\
                                                                                      & 3                             & \multicolumn{1}{c|}{76.1} & 82.7 & \textbf{82.8}  &                      \\
                                                                                      & 4                             & \multicolumn{1}{c|}{69.1} & 75.8 & \textbf{76.0}  &                      \\
                                                                                      & mean                          & \multicolumn{1}{c|}{78.9} & 83.7 & \textbf{84.5}  &                      \\ \cline{1-5}
\multirow{5}{*}{\begin{tabular}[c]{@{}c@{}}A new room\end{tabular}}                                                       & 1                             & \multicolumn{1}{c|}{80.9}     & 89.9 & \textbf{92.3}  &                      \\
                                                                                      & 2                             & \multicolumn{1}{c|}{79.3}     & 92.4 & \textbf{93.0}  & \multicolumn{1}{c}{} \\
                                                                                      & 3                             & \multicolumn{1}{c|}{73.0}     & \textbf{86.5} & 85.6  & \multicolumn{1}{c}{} \\
                                                                                      & 4                             & \multicolumn{1}{c|}{58.7}     & 70.7 & \textbf{87.6}  & \multicolumn{1}{c}{} \\
                                                                                      & mean                          & \multicolumn{1}{c|}{72.6} & 86.4 & \textbf{87.6}  & \multicolumn{1}{c}{} \\ \cline{1-5}
\end{tabular}
}
\caption{Results of PCKh\textsubscript{50} for multiple people in test scenario (A) changing antenna locations in the same room and (C) in untrained environment.}
\label{table:different room performance}
\end{table}

\begin{table}[ht]
\centering\resizebox{0.8\columnwidth}{!}{
\renewcommand{\arraystretch}{1.2}
\begin{tabular}{cc|cccl}
\cline{1-5}
\multirow{2}{*}{Test set}                                                             & \multirow{2}{*}{\# of person} & \multicolumn{3}{c}{Model}                &                      \\ \cline{3-5}
                                                                                      &                               & \multicolumn{1}{c|}{RPET} & OS & SOS &                      \\ \cline{1-5}
\multirow{5}{*}{\begin{tabular}[c]{@{}c@{}}Obstacle\\ Cotton Fabric\end{tabular}}  & 1                             & \multicolumn{1}{c|}{76.4}   & 81.3 & \textbf{83.3}  &                      \\
                                                                                      & 2                             & \multicolumn{1}{c|}{75.1}     & 82.2 & \textbf{85.7}  &                      \\
                                                                                      & 3                             & \multicolumn{1}{c|}{70.7}     & 77.4 & \textbf{79.1}  &                      \\
                                                                                      & 4                             & \multicolumn{1}{c|}{66.8}     & 72.8 & \textbf{77.4}  &                      \\
                                                                                      & mean                          & \multicolumn{1}{c|}{74.6} & 78.4 & \textbf{82.1}  &                      \\ \cline{1-5}
\multirow{5}{*}{\begin{tabular}[c]{@{}c@{}}Obstacle\\ Foam box\end{tabular}} & 1                             & \multicolumn{1}{c|}{82.8}     & 88.3 & \textbf{90.0}  &                      \\
                                                                                      & 2                             & \multicolumn{1}{c|}{78.6}     & 84.9 & \textbf{90.6}  &                      \\
                                                                                      & 3                             & \multicolumn{1}{c|}{74.1}     & 79.1 & \textbf{80.1}  &                      \\
                                                                                      & 4                             & \multicolumn{1}{c|}{63.6}     & 68.3 & \textbf{71.3}  &                      \\
                                                                                      & mean                          & \multicolumn{1}{c|}{76.6} & 80.1 & \textbf{83.0}  &                      \\ \cline{1-5}
\end{tabular}
}
\caption{Results of PCKh\textsubscript{50} for multiple people in test scenario (B) placing obstacles in front of the radar antennas. There were two obstacles to the test data set. One is cotton fabric and the other is a foam box.}
\label{table:obstacle performance}
\end{table}

\subsection{Impact of Self-Supervised Learning} \label{section-ssl}

We demonstrate in Table \ref{tab:table-jointresults} that the performance of the proposed Self-Supervised Learning (SSL) pre-trained and MPPE fine-tuned SOS model surpasses that of the OS model. Specifically, SOS outperforms OS in performance across all body keypoints. Notably, for small body keypoints such as the wrist, both the proposed OS and SOS models show significant performance improvements compared to RPET.

Table~\ref{table:different room performance}. shows performance relative to the number of people in the observed location, with SOS generally performing better even in entirely new locations (C). Table~\ref{table:obstacle performance} presents results with an obstacle placed in front of the radar antenna (B), which weakens the received signal strength, causing variations in data distribution and degrading model performance. In these challenging test scenario, the OS model outperforms RPET, and SOS demonstrates the best performance. These results highlight the generalization capabilities of our proposed models, despite being untrained on such scenarios, and validate the effectiveness of our newly proposed SSL technique.

\subsection{Ablation Study of Self-Supervised Learning}

\begin{table}[]
\centering\resizebox{0.8\columnwidth}{!}{\tiny
\renewcommand{\arraystretch}{1.3}
\begin{tabular}{c|c}
\hline
 Self-Supervised Methods & PCKh\textsubscript{50} \\ 
 \hline 
 SOS without Cross-Attention &   81.9 (-1.8)   \\ 
 
\begin{tabular}[c]{@{}c@{}}SOS with Raw-Signal Reconstruction\end{tabular} & 83.9 (+0.2)  \\ 

Proposed SOS & \textbf{84.5 (+0.8)}   \\ 
\hline
\end{tabular}
}
\caption{Our self-supervised learning approach utilizes cross-attention with a unmasked subgroup and reconstruction at the latent level, which clearly outperforms both the method using only self-attention for latent-level restoration without cross-attention and the method that reconstruct data at the raw signal level.}

\label{table-SSL-ablation}
\end{table}
Our proposed self-supervised learning features two key characteristics. First, it uses cross-attention to propagate information from one unmasked subgroup to the masked other subgroups. Second, it restores the masked parts at the representation latent level. As an ablation test, we do not apply cross-attention. Instead, we apply masking to all subgroups and attempt latent-level restoration using only self-attention between the unmasked patches, similar to data2vec \citep{baevski2023efficient}. Another ablation study is to retain the cross-attention technique but restore at the raw-signal level rather than the latent level. This can be considered akin to pixel-level restoration in the image domain \citep{he2022masked, gupta2023siamese}. Performance differences according to these techniques can be observed in Table~\ref{table-SSL-ablation}. It is evident that the application of cross-attention significantly impacts performance, and the proposed SOS performance shows superior results compared to the raw signal-level restoration method.

\subsection{Prediction with Only One Subgroup}
\begin{table}[]
{\tiny
\centering\resizebox{0.75\columnwidth}{!}{
\renewcommand{\arraystretch}{1.1}
\begin{tabular}{c|cccc}
\hline
\multirow{2}{*}{Test set}                                                      & \multicolumn{3}{c}{Model}                      \\ \cline{2-4} 
                                                                               & OS$^\dagger$     & OS*  & SOS* \\ \hline
\begin{tabular}[c]{@{}c@{}}Different locations\\ in the same room\end{tabular} & 76.7  & 77.8 & \textbf{78.9} \\ \hline
\begin{tabular}[c]{@{}c@{}}Obstacle\\ Cotton Fabric\end{tabular}               & 74.7  & 73.5 & \textbf{75.0} \\ \hline
\begin{tabular}[c]{@{}c@{}}Obstacle\\ Foam box\end{tabular}                    & 76.5  & 76.6 & \textbf{77.8} \\ \hline
A new room                                                                     & 78.0  & 82.6 & \textbf{83.4} \\ \hline
\end{tabular}
}
\caption{Results of PCKh\textsubscript{50} for using only one subgroup. * denotes using all subgroup for training and using only one subgroup for inference. $\dagger$ denotes using one subgroup for both training and inference.}
\label{table-One subgroup}}
\end{table}

\begin{figure*}[!h]
\centering
\includegraphics[trim= 100 140 160 65, clip, width=1.0\textwidth]{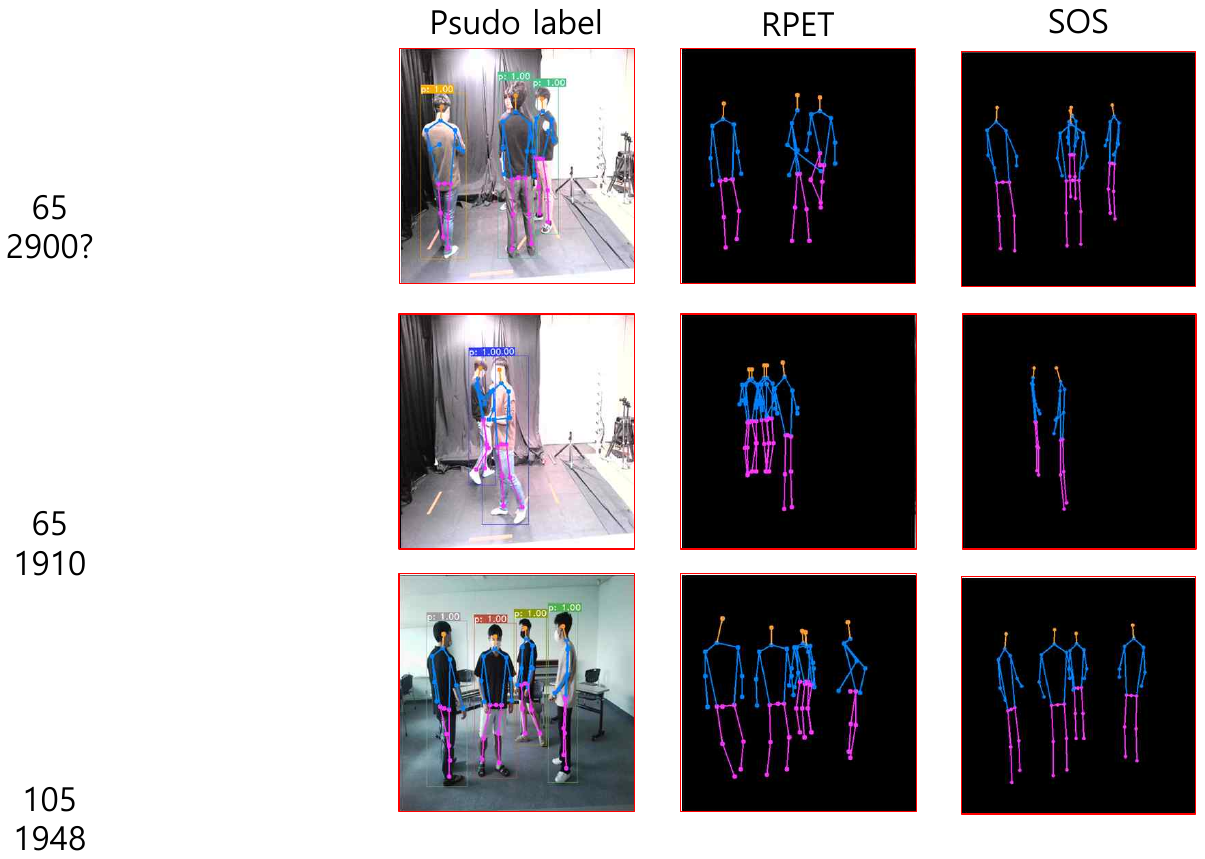}
\caption{Comparison inference results of RPET and SOS. The main improvements are in the reduction of pose collapse, handling of side-view scenes, and detection of the number of people. First row is four people in training environment. Second row is two people in training environment. The second row shows a scene where two people are seen in a side view within the training environment. The third row is a situation with four people in a new environment featuring structures like desks}
\label{fig-inference}
\end{figure*}

In this section, we evaluate the performance of our trained model when using only one subgroup is used as input. This scenario is equivalent to using a model trained with an 8x8 antenna radar to evaluate MPPE with signals from a radar equipped with only a 4x4 antenna array. Our model structure can operate without modifications, even with fewer subgroups used as inputs. The model, OS$^\dagger$, in Table ~\ref{table-One subgroup} was trained using only one subgroup. Since only one subgroup is used, the subgroup-level SSL (proposed in this paper) could not be applied, and thus only OS$^\dagger$ results are presented. OS* and SOS* were trained using all four subgroups, but evaluated using only one subgroup input. OS* shows similar or generally improved performance compared to OS$^\dagger$, and SOS* shows consistent performance improvements up to 5.3 compared to OS$^\dagger$. This is particularly meaningful as it can minimize performance degradation even when the number of subgroups used as inputs is reduced due to hardware limitations or defects in real deployments.

\subsection{Qualitative Evaluation Comparison}

This section presents a qualitative evaluation. As shown in Fig ~\ref{fig-inference}, RPET~\cite{kim2024learning} had difficulties in estimating poses in overlapping scenarios. Specifically, it struggled with predicting scenes where people are standing sideways rather than facing forward, and with pose estimation when evaluated in new environments. In contrast, our proposed SOS model demonstrates improvements in these situations.

%% file: 7.Conclusion.tex
\section{Conclusion}

This paper introduces an efficient one-stage model for Multi-Person Pose Estimation (MPPE) using raw RF signals. The employment of a shared single-layer CNN and multi-head attention to embed sub-grouped RF signals has proven to be a highly effective strategy. This method not only reduces the model complexity (drastically reduced the number of parameters by 98\%) but also leverages the representation of RF signals to improve MPPE performance. We also proposed a novel self-supervised learning approach which further improves the representation learning and generalization performance of raw RF signal-based MPPE. Finally, by making all source codes and datasets used in this work publicly available, we hope to facilitate ongoing progress and innovation in this field.

%% file: main.bbl

\begin{thebibliography}{30}


\ifx \showCODEN    \undefined \def \showCODEN     #1{\unskip}     \fi
\ifx \showDOI      \undefined \def \showDOI       #1{#1}\fi
\ifx \showISBNx    \undefined \def \showISBNx     #1{\unskip}     \fi
\ifx \showISBNxiii \undefined \def \showISBNxiii  #1{\unskip}     \fi
\ifx \showISSN     \undefined \def \showISSN      #1{\unskip}     \fi
\ifx \showLCCN     \undefined \def \showLCCN      #1{\unskip}     \fi
\ifx \shownote     \undefined \def \shownote      #1{#1}          \fi
\ifx \showarticletitle \undefined \def \showarticletitle #1{#1}   \fi
\ifx \showURL      \undefined \def \showURL       {\relax}        \fi
\providecommand\bibfield[2]{#2}
\providecommand\bibinfo[2]{#2}
\providecommand\natexlab[1]{#1}
\providecommand\showeprint[2][]{arXiv:#2}

\bibitem[Adib et~al\mbox{.}(2014)]%
        {adib20143d}
\bibfield{author}{\bibinfo{person}{Fadel Adib}, \bibinfo{person}{Zach Kabelac}, \bibinfo{person}{Dina Katabi}, {and} \bibinfo{person}{Robert~C Miller}.} \bibinfo{year}{2014}\natexlab{}.
\newblock \showarticletitle{3D tracking via body radio reflections}. In \bibinfo{booktitle}{\emph{11th USENIX Symposium on Networked Systems Design and Implementation (NSDI 14)}}. \bibinfo{pages}{317--329}.
\newblock


\bibitem[Baevski et~al\mbox{.}(2023)]%
        {baevski2023efficient}
\bibfield{author}{\bibinfo{person}{Alexei Baevski}, \bibinfo{person}{Arun Babu}, \bibinfo{person}{Wei-Ning Hsu}, {and} \bibinfo{person}{Michael Auli}.} \bibinfo{year}{2023}\natexlab{}.
\newblock \showarticletitle{Efficient self-supervised learning with contextualized target representations for vision, speech and language}. In \bibinfo{booktitle}{\emph{International Conference on Machine Learning}}. PMLR, \bibinfo{pages}{1416--1429}.
\newblock


\bibitem[Baevski et~al\mbox{.}(2022)]%
        {baevski2022data2vec}
\bibfield{author}{\bibinfo{person}{Alexei Baevski}, \bibinfo{person}{Wei-Ning Hsu}, \bibinfo{person}{Qiantong Xu}, \bibinfo{person}{Arun Babu}, \bibinfo{person}{Jiatao Gu}, {and} \bibinfo{person}{Michael Auli}.} \bibinfo{year}{2022}\natexlab{}.
\newblock \showarticletitle{Data2vec: A general framework for self-supervised learning in speech, vision and language}. In \bibinfo{booktitle}{\emph{International Conference on Machine Learning}}. PMLR, \bibinfo{pages}{1298--1312}.
\newblock


\bibitem[Bodla et~al\mbox{.}(2017)]%
        {bodla2017soft}
\bibfield{author}{\bibinfo{person}{Navaneeth Bodla}, \bibinfo{person}{Bharat Singh}, \bibinfo{person}{Rama Chellappa}, {and} \bibinfo{person}{Larry~S Davis}.} \bibinfo{year}{2017}\natexlab{}.
\newblock \showarticletitle{Soft-NMS--improving object detection with one line of code}. In \bibinfo{booktitle}{\emph{Proceedings of the IEEE international conference on computer vision}}. \bibinfo{pages}{5561--5569}.
\newblock


\bibitem[Cao et~al\mbox{.}(2017)]%
        {cao2017realtime}
\bibfield{author}{\bibinfo{person}{Zhe Cao}, \bibinfo{person}{Tomas Simon}, \bibinfo{person}{Shih-En Wei}, {and} \bibinfo{person}{Yaser Sheikh}.} \bibinfo{year}{2017}\natexlab{}.
\newblock \showarticletitle{Realtime multi-person 2d pose estimation using part affinity fields}. In \bibinfo{booktitle}{\emph{Proceedings of the IEEE conference on computer vision and pattern recognition}}. \bibinfo{pages}{7291--7299}.
\newblock


\bibitem[Carion et~al\mbox{.}(2020)]%
        {carion2020end}
\bibfield{author}{\bibinfo{person}{Nicolas Carion}, \bibinfo{person}{Francisco Massa}, \bibinfo{person}{Gabriel Synnaeve}, \bibinfo{person}{Nicolas Usunier}, \bibinfo{person}{Alexander Kirillov}, {and} \bibinfo{person}{Sergey Zagoruyko}.} \bibinfo{year}{2020}\natexlab{}.
\newblock \showarticletitle{End-to-end object detection with transformers}. In \bibinfo{booktitle}{\emph{European conference on computer vision}}. Springer, \bibinfo{pages}{213--229}.
\newblock


\bibitem[Chen and He(2021)]%
        {chen2021exploring}
\bibfield{author}{\bibinfo{person}{Xinlei Chen} {and} \bibinfo{person}{Kaiming He}.} \bibinfo{year}{2021}\natexlab{}.
\newblock \showarticletitle{Exploring simple siamese representation learning}. In \bibinfo{booktitle}{\emph{Proceedings of the IEEE/CVF conference on computer vision and pattern recognition}}. \bibinfo{pages}{15750--15758}.
\newblock


\bibitem[Cheng et~al\mbox{.}(2020)]%
        {cheng2020higherhrnet}
\bibfield{author}{\bibinfo{person}{Bowen Cheng}, \bibinfo{person}{Bin Xiao}, \bibinfo{person}{Jingdong Wang}, \bibinfo{person}{Honghui Shi}, \bibinfo{person}{Thomas~S Huang}, {and} \bibinfo{person}{Lei Zhang}.} \bibinfo{year}{2020}\natexlab{}.
\newblock \showarticletitle{Higherhrnet: Scale-aware representation learning for bottom-up human pose estimation}. In \bibinfo{booktitle}{\emph{Proceedings of the IEEE/CVF conference on computer vision and pattern recognition}}. \bibinfo{pages}{5386--5395}.
\newblock


\bibitem[Dosovitskiy et~al\mbox{.}(2020)]%
        {dosovitskiy2020image}
\bibfield{author}{\bibinfo{person}{Alexey Dosovitskiy}, \bibinfo{person}{Lucas Beyer}, \bibinfo{person}{Alexander Kolesnikov}, \bibinfo{person}{Dirk Weissenborn}, \bibinfo{person}{Xiaohua Zhai}, \bibinfo{person}{Thomas Unterthiner}, \bibinfo{person}{Mostafa Dehghani}, \bibinfo{person}{Matthias Minderer}, \bibinfo{person}{Georg Heigold}, \bibinfo{person}{Sylvain Gelly}, {et~al\mbox{.}}} \bibinfo{year}{2020}\natexlab{}.
\newblock \showarticletitle{An image is worth 16x16 words: Transformers for image recognition at scale}.
\newblock \bibinfo{journal}{\emph{arXiv preprint arXiv:2010.11929}} (\bibinfo{year}{2020}).
\newblock


\bibitem[Gupta et~al\mbox{.}(2023)]%
        {gupta2023siamese}
\bibfield{author}{\bibinfo{person}{Agrim Gupta}, \bibinfo{person}{Jiajun Wu}, \bibinfo{person}{Jia Deng}, {and} \bibinfo{person}{Li Fei-Fei}.} \bibinfo{year}{2023}\natexlab{}.
\newblock \showarticletitle{Siamese Masked Autoencoders}.
\newblock \bibinfo{journal}{\emph{arXiv preprint arXiv:2305.14344}} (\bibinfo{year}{2023}).
\newblock


\bibitem[He et~al\mbox{.}(2022)]%
        {he2022masked}
\bibfield{author}{\bibinfo{person}{Kaiming He}, \bibinfo{person}{Xinlei Chen}, \bibinfo{person}{Saining Xie}, \bibinfo{person}{Yanghao Li}, \bibinfo{person}{Piotr Doll{\'a}r}, {and} \bibinfo{person}{Ross Girshick}.} \bibinfo{year}{2022}\natexlab{}.
\newblock \showarticletitle{Masked autoencoders are scalable vision learners}. In \bibinfo{booktitle}{\emph{Proceedings of the IEEE/CVF conference on computer vision and pattern recognition}}. \bibinfo{pages}{16000--16009}.
\newblock


\bibitem[He et~al\mbox{.}(2017)]%
        {he2017mask}
\bibfield{author}{\bibinfo{person}{Kaiming He}, \bibinfo{person}{Georgia Gkioxari}, \bibinfo{person}{Piotr Doll{\'a}r}, {and} \bibinfo{person}{Ross Girshick}.} \bibinfo{year}{2017}\natexlab{}.
\newblock \showarticletitle{Mask r-cnn}. In \bibinfo{booktitle}{\emph{Proceedings of the IEEE international conference on computer vision}}. \bibinfo{pages}{2961--2969}.
\newblock


\bibitem[He et~al\mbox{.}(2016)]%
        {he2016deep}
\bibfield{author}{\bibinfo{person}{Kaiming He}, \bibinfo{person}{Xiangyu Zhang}, \bibinfo{person}{Shaoqing Ren}, {and} \bibinfo{person}{Jian Sun}.} \bibinfo{year}{2016}\natexlab{}.
\newblock \showarticletitle{Deep residual learning for image recognition}. In \bibinfo{booktitle}{\emph{Proceedings of the IEEE conference on computer vision and pattern recognition}}. \bibinfo{pages}{770--778}.
\newblock


\bibitem[Kim et~al\mbox{.}(2024)]%
        {kim2024learning}
\bibfield{author}{\bibinfo{person}{Seunghyun Kim}, \bibinfo{person}{Seunghwan Shin}, \bibinfo{person}{Sangwon Lee}, \bibinfo{person}{Kaewon Choi}, {and} \bibinfo{person}{Yusung Kim}.} \bibinfo{year}{2024}\natexlab{}.
\newblock \showarticletitle{Learning Visual Clue for UWB-based multi-person pose estimation}.
\newblock \bibinfo{journal}{\emph{Knowledge-Based Systems}}  \bibinfo{volume}{284} (\bibinfo{year}{2024}), \bibinfo{pages}{111289}.
\newblock


\bibitem[Krizhevsky et~al\mbox{.}(2012)]%
        {krizhevsky2012imagenet}
\bibfield{author}{\bibinfo{person}{Alex Krizhevsky}, \bibinfo{person}{Ilya Sutskever}, {and} \bibinfo{person}{Geoffrey~E Hinton}.} \bibinfo{year}{2012}\natexlab{}.
\newblock \showarticletitle{Imagenet classification with deep convolutional neural networks}.
\newblock \bibinfo{journal}{\emph{Advances in neural information processing systems}}  \bibinfo{volume}{25} (\bibinfo{year}{2012}).
\newblock


\bibitem[Li et~al\mbox{.}(2021)]%
        {li2021pose}
\bibfield{author}{\bibinfo{person}{Ke Li}, \bibinfo{person}{Shijie Wang}, \bibinfo{person}{Xiang Zhang}, \bibinfo{person}{Yifan Xu}, \bibinfo{person}{Weijian Xu}, {and} \bibinfo{person}{Zhuowen Tu}.} \bibinfo{year}{2021}\natexlab{}.
\newblock \showarticletitle{Pose recognition with cascade transformers}. In \bibinfo{booktitle}{\emph{Proceedings of the IEEE/CVF conference on computer vision and pattern recognition}}. \bibinfo{pages}{1944--1953}.
\newblock


\bibitem[Liu et~al\mbox{.}(2023)]%
        {liu2023group}
\bibfield{author}{\bibinfo{person}{Huan Liu}, \bibinfo{person}{Qiang Chen}, \bibinfo{person}{Zichang Tan}, \bibinfo{person}{Jiang-Jiang Liu}, \bibinfo{person}{Jian Wang}, \bibinfo{person}{Xiangbo Su}, \bibinfo{person}{Xiaolong Li}, \bibinfo{person}{Kun Yao}, \bibinfo{person}{Junyu Han}, \bibinfo{person}{Errui Ding}, {et~al\mbox{.}}} \bibinfo{year}{2023}\natexlab{}.
\newblock \showarticletitle{Group Pose: A Simple Baseline for End-to-End Multi-person Pose Estimation}. In \bibinfo{booktitle}{\emph{Proceedings of the IEEE/CVF International Conference on Computer Vision}}. \bibinfo{pages}{15029--15038}.
\newblock


\bibitem[Liu et~al\mbox{.}(2021)]%
        {liu2021swin}
\bibfield{author}{\bibinfo{person}{Ze Liu}, \bibinfo{person}{Yutong Lin}, \bibinfo{person}{Yue Cao}, \bibinfo{person}{Han Hu}, \bibinfo{person}{Yixuan Wei}, \bibinfo{person}{Zheng Zhang}, \bibinfo{person}{Stephen Lin}, {and} \bibinfo{person}{Baining Guo}.} \bibinfo{year}{2021}\natexlab{}.
\newblock \showarticletitle{Swin transformer: Hierarchical vision transformer using shifted windows}. In \bibinfo{booktitle}{\emph{Proceedings of the IEEE/CVF international conference on computer vision}}. \bibinfo{pages}{10012--10022}.
\newblock


\bibitem[Liu et~al\mbox{.}(2022)]%
        {liu2022convnet}
\bibfield{author}{\bibinfo{person}{Zhuang Liu}, \bibinfo{person}{Hanzi Mao}, \bibinfo{person}{Chao-Yuan Wu}, \bibinfo{person}{Christoph Feichtenhofer}, \bibinfo{person}{Trevor Darrell}, {and} \bibinfo{person}{Saining Xie}.} \bibinfo{year}{2022}\natexlab{}.
\newblock \showarticletitle{A convnet for the 2020s}. In \bibinfo{booktitle}{\emph{Proceedings of the IEEE/CVF conference on computer vision and pattern recognition}}. \bibinfo{pages}{11976--11986}.
\newblock


\bibitem[Newell et~al\mbox{.}(2017)]%
        {newell2017associative}
\bibfield{author}{\bibinfo{person}{Alejandro Newell}, \bibinfo{person}{Zhiao Huang}, {and} \bibinfo{person}{Jia Deng}.} \bibinfo{year}{2017}\natexlab{}.
\newblock \showarticletitle{Associative embedding: End-to-end learning for joint detection and grouping}.
\newblock \bibinfo{journal}{\emph{Advances in neural information processing systems}}  \bibinfo{volume}{30} (\bibinfo{year}{2017}).
\newblock


\bibitem[Shi et~al\mbox{.}(2022)]%
        {shi2022end}
\bibfield{author}{\bibinfo{person}{Dahu Shi}, \bibinfo{person}{Xing Wei}, \bibinfo{person}{Liangqi Li}, \bibinfo{person}{Ye Ren}, {and} \bibinfo{person}{Wenming Tan}.} \bibinfo{year}{2022}\natexlab{}.
\newblock \showarticletitle{End-to-end multi-person pose estimation with transformers}. In \bibinfo{booktitle}{\emph{Proceedings of the IEEE/CVF Conference on Computer Vision and Pattern Recognition}}. \bibinfo{pages}{11069--11078}.
\newblock


\bibitem[Simonyan and Zisserman(2014)]%
        {simonyan2014very}
\bibfield{author}{\bibinfo{person}{Karen Simonyan} {and} \bibinfo{person}{Andrew Zisserman}.} \bibinfo{year}{2014}\natexlab{}.
\newblock \showarticletitle{Very deep convolutional networks for large-scale image recognition}.
\newblock \bibinfo{journal}{\emph{arXiv preprint arXiv:1409.1556}} (\bibinfo{year}{2014}).
\newblock


\bibitem[Sun et~al\mbox{.}(2019)]%
        {sun2019deep}
\bibfield{author}{\bibinfo{person}{Ke Sun}, \bibinfo{person}{Bin Xiao}, \bibinfo{person}{Dong Liu}, {and} \bibinfo{person}{Jingdong Wang}.} \bibinfo{year}{2019}\natexlab{}.
\newblock \showarticletitle{Deep high-resolution representation learning for human pose estimation}. In \bibinfo{booktitle}{\emph{Proceedings of the IEEE/CVF conference on computer vision and pattern recognition}}. \bibinfo{pages}{5693--5703}.
\newblock


\bibitem[Taylor and Wisland(2017)]%
        {taylor:17}
\bibfield{author}{\bibinfo{person}{James~D. Taylor} {and} \bibinfo{person}{Dag~T. Wisland}.} \bibinfo{year}{2017}\natexlab{}.
\newblock \showarticletitle{Novelda Nanoscale Impulse Radar}.
\newblock


\bibitem[Vaswani et~al\mbox{.}(2017)]%
        {vaswani2017attention}
\bibfield{author}{\bibinfo{person}{Ashish Vaswani}, \bibinfo{person}{Noam Shazeer}, \bibinfo{person}{Niki Parmar}, \bibinfo{person}{Jakob Uszkoreit}, \bibinfo{person}{Llion Jones}, \bibinfo{person}{Aidan~N Gomez}, \bibinfo{person}{{\L}ukasz Kaiser}, {and} \bibinfo{person}{Illia Polosukhin}.} \bibinfo{year}{2017}\natexlab{}.
\newblock \showarticletitle{Attention is all you need}.
\newblock \bibinfo{journal}{\emph{Advances in neural information processing systems}}  \bibinfo{volume}{30} (\bibinfo{year}{2017}).
\newblock


\bibitem[Wang et~al\mbox{.}(2019)]%
        {wang2019person}
\bibfield{author}{\bibinfo{person}{Fei Wang}, \bibinfo{person}{Sanping Zhou}, \bibinfo{person}{Stanislav Panev}, \bibinfo{person}{Jinsong Han}, {and} \bibinfo{person}{Dong Huang}.} \bibinfo{year}{2019}\natexlab{}.
\newblock \showarticletitle{Person-in-WiFi: Fine-grained person perception using WiFi}. In \bibinfo{booktitle}{\emph{Proceedings of the IEEE/CVF International Conference on Computer Vision}}. \bibinfo{pages}{5452--5461}.
\newblock


\bibitem[Xiao et~al\mbox{.}(2018)]%
        {xiao2018simple}
\bibfield{author}{\bibinfo{person}{Bin Xiao}, \bibinfo{person}{Haiping Wu}, {and} \bibinfo{person}{Yichen Wei}.} \bibinfo{year}{2018}\natexlab{}.
\newblock \showarticletitle{Simple baselines for human pose estimation and tracking}. In \bibinfo{booktitle}{\emph{Proceedings of the European conference on computer vision (ECCV)}}. \bibinfo{pages}{466--481}.
\newblock


\bibitem[Xu et~al\mbox{.}(2022)]%
        {xu2022vitpose}
\bibfield{author}{\bibinfo{person}{Yufei Xu}, \bibinfo{person}{Jing Zhang}, \bibinfo{person}{Qiming Zhang}, {and} \bibinfo{person}{Dacheng Tao}.} \bibinfo{year}{2022}\natexlab{}.
\newblock \showarticletitle{Vitpose: Simple vision transformer baselines for human pose estimation}.
\newblock \bibinfo{journal}{\emph{Advances in Neural Information Processing Systems}}  \bibinfo{volume}{35} (\bibinfo{year}{2022}), \bibinfo{pages}{38571--38584}.
\newblock


\bibitem[Zhao et~al\mbox{.}(2018)]%
        {zhao2018through}
\bibfield{author}{\bibinfo{person}{Mingmin Zhao}, \bibinfo{person}{Tianhong Li}, \bibinfo{person}{Mohammad Abu~Alsheikh}, \bibinfo{person}{Yonglong Tian}, \bibinfo{person}{Hang Zhao}, \bibinfo{person}{Antonio Torralba}, {and} \bibinfo{person}{Dina Katabi}.} \bibinfo{year}{2018}\natexlab{}.
\newblock \showarticletitle{Through-wall human pose estimation using radio signals}. In \bibinfo{booktitle}{\emph{Proceedings of the IEEE conference on computer vision and pattern recognition}}. \bibinfo{pages}{7356--7365}.
\newblock


\bibitem[Zheng et~al\mbox{.}(2022)]%
        {zheng2022recovering}
\bibfield{author}{\bibinfo{person}{Zhijie Zheng}, \bibinfo{person}{Jun Pan}, \bibinfo{person}{Zhikang Ni}, \bibinfo{person}{Cheng Shi}, \bibinfo{person}{Diankun Zhang}, \bibinfo{person}{Xiaojun Liu}, {and} \bibinfo{person}{Guangyou Fang}.} \bibinfo{year}{2022}\natexlab{}.
\newblock \showarticletitle{Recovering human pose and shape from through-the-wall radar images}.
\newblock \bibinfo{journal}{\emph{IEEE Transactions on Geoscience and Remote Sensing}}  \bibinfo{volume}{60} (\bibinfo{year}{2022}), \bibinfo{pages}{1--15}.
\newblock


\end{thebibliography}
